\documentclass[sn-mathphys,Numbered]{sn-jnl}

\usepackage{graphicx}%
\usepackage{multirow}%
\usepackage{amsmath,amssymb,amsfonts}%
\usepackage{amsthm}%
\usepackage{mathrsfs}%
\usepackage[title]{appendix}%
\usepackage{xcolor}%
\usepackage{textcomp}%
\usepackage{manyfoot}%
\usepackage{booktabs}%
\usepackage{algorithm}%
\usepackage{algorithmicx}%
\usepackage{algpseudocode}%
\usepackage{listings}%

\theoremstyle{thmstyleone}%
\theoremstyle{thmstyletwo}%

\theoremstyle{thmstylethree}%

\begin{document}

\title[CellAgent]{CellAgent: An LLM-driven Multi-Agent Framework for Automated Single-cell Data Analysis}

\author[1]{\fnm{Yihang} \sur{Xiao}}\equalcont{These authors contributed equally to this work.}
\author[2]{\fnm{Jinyi} \sur{Liu}}\equalcont{These authors contributed equally to this work.}
\author[2]{\fnm{Yan} \sur{Zheng}}\equalcont{These authors contributed equally to this work.}
\author[1]{\fnm{Xiaohan} \sur{Xie}}\equalcont{These authors contributed equally to this work.}
\author*[2]{\fnm{Jianye} \sur{Hao}}\email{jianye.hao@tju.edu.cn}
\author[2]{\fnm{Mingzhi} \sur{Li}}
\author[2]{\fnm{Ruitao} \sur{Wang}}
\author[2]{\fnm{Fei} \sur{Ni}}
\author[2]{\fnm{Yuxiao} \sur{Li}}
\author[1]{\fnm{Jintian} \sur{Luo}}
\author[1]{\fnm{Shaoqing} \sur{Jiao}}
\author*[1,3]{\fnm{Jiajie} \sur{Peng}}\email{jiajiepeng@nwpu.edu.cn}

\affil*[1]{\orgdiv{AI for Science Interdisciplinary Research Center}, \orgname{School of Computer Science, Northwestern Polytechnical University}, \orgaddress{\street{No.1 Dongxiang Road}, \city{Xi'an}, \country{China}}}
\affil[2]{\orgdiv{College of Intelligence and Computing}, \orgname{Tianjin University}, \orgaddress{\street{No.92 Weijin Road}, \city{Tianjin}, \country{China}}}

\affil*[3]{\orgdiv{Key Laboratory of Big Data Storage and Management, Northwestern Polytechnical University, Ministry of Industry and Information Technology}, \orgaddress{\street{No.1 Dongxiang Road}, \city{Xi'an}, \country{China}}}

\abstract{Single-cell RNA sequencing (scRNA-seq) data analysis is crucial for biological research, as it enables the precise characterization of cellular heterogeneity. However, manual manipulation of various tools to achieve desired outcomes can be labor-intensive for researchers. 
To address this, we introduce CellAgent (\url{http://cell.agent4science.cn/}), an LLM-driven multi-agent framework, specifically designed for the automatic processing and execution of scRNA-seq data analysis tasks, providing high-quality results with no human intervention.
Firstly, to adapt general LLMs to the biological field, CellAgent constructs LLM-driven biological expert roles—planner, executor, and evaluator—each with specific responsibilities.
Then, CellAgent introduces a hierarchical decision-making mechanism to coordinate these biological experts, effectively driving the planning and step-by-step execution of complex data analysis tasks.
Furthermore, we propose a self-iterative optimization mechanism, enabling CellAgent to autonomously evaluate and optimize solutions, thereby guaranteeing output quality.
We evaluate CellAgent on a comprehensive benchmark dataset encompassing dozens of tissues and hundreds of distinct cell types. Evaluation results consistently show that CellAgent effectively identifies the most suitable tools and hyperparameters for single-cell analysis tasks, achieving optimal performance.
This automated framework dramatically reduces the workload for science data analyses, bringing us into the ``Agent for Science" era.

}

\maketitle

\section{Introduction}
Single-cell RNA sequencing (scRNA-seq) techniques have transformed molecular biology by allowing for the analysis of transcriptome profiles with unprecedented scale and precision~\cite{scpipeline}. Such advances have driven large-scale innovations in computational methods, resulting in over 1400 tools currently available for analyzing scRNA-seq data from various perspectives ~\cite{singlecelldevelopment}. 
These tools, including computational frameworks and software libraries~\cite{bioconductor, seurat, scanpy}, coupled with method benchmarks and best practice workflows~\cite{pipeline_sc,bestpratice}, lead to groundbreaking discoveries in biology~\cite{eraslan2022single,sikkema2023integrated}.  

However, analyzing scRNA-seq data entails considerable complexity and demands specialized knowledge and expertise.
The steps involved may include preprocessing, batch correction, clustering, marker gene visualization, cell type annotation, and trajectory inference, among others.
To complete these steps, researchers must carefully execute a series of corresponding tools while configuring appropriate hyperparameters and models tailored to the specific characteristics of the biological data~\cite{bestpratice}.
This requires researchers to possess not only advanced programming skills but also a strong biological background, further increasing the cost of conducting single-cell analysis tasks. Therefore, there is an urgent need for an intelligent tool capable of automatically assembling and executing existing tools to generate analysis results for scRNA-seq data. Such an automated tool could significantly reduce the technical barriers in the field of biology and remove obstacles in data analysis for biological scientists.

Recently, with the remarkable capabilities demonstrated by large language models (LLMs)~\cite{GPT4,palm2,Llama2}, there has been growing research interest in LLM-driven autonomous AI agents capable of automatically handling tasks, such as MetaGPT~\cite{hong2023metagpt} for software programming and Microsoft Copilot~\cite{bingcopilot} for office suite assistance. These AI agents typically employ LLMs as the core of the agent's brain, supplemented by necessary mechanisms such as memory and tools to expand their capabilities~\cite{zhou2023agents,agentaisurvey1,agentaisurvey2}. Inspired by these advancements, a novel question arises: \textit{How to employ LLMs to design a biologically proficient agent framework for automating single-cell data analysis tasks?}

Applying LLMs directly to complex scRNA-seq data analysis poses significant challenges. General LLMs lack comprehensive and accurate knowledge of specialized biological tools and concepts, leading to outputs that may lack reasonable biological significance. Moreover, LLMs struggle to comprehend the lengthy context involved in multi-step tasks, easily losing crucial information. Nevertheless, LLMs are unaware of these deficiencies, resulting in unreliable outcomes. Therefore, to develop specialized approaches that effectively integrate LLMs' strengths with the specific requirements of scRNA-seq data analysis is crucial, for achieving efficient, professional, and automated task execution.

In this paper, to address the aforementioned challenges, we propose a specialized, zero-code CellAgent, an LLM-driven multi-agent collaborative framework for scRNA-seq data analysis. CellAgent can directly comprehend natural language task descriptions, completing complex tasks with high quality through effective collaboration, autonomously.
Firstly, considering the limitations of LLMs in understanding biological expertise, we manually gather a few crucial expert experiences and tools. Three types of biological expert roles driven by LLM are established: Planner, Executor, and Evaluator. To enhance the stability of the execution process and avoid handling lengthy steps in a single pass, CellAgent introduces a hierarchical decision-making mechanism, with upper-level task planning via Planner, and lower-level task execution via Executor. Furthermore, to guarantee high-quality solution outputs from CellAgent, we propose a self-iterative optimization mechanism, encouraging Executors to autonomously optimize the planning process by incorporating automated evaluation results and accounting for potential code execution exceptions. The Evaluator plays a crucial role in driving the iterative optimization process, possessing the ability to assess the quality of task solutions from a biological expert's perspective.
Through the effective collaboration of these biological experts, CellAgent finally possesses robust capabilities for automated scRNA-seq data analysis.
Without requiring human intervention, CellAgent enables the end-to-end automatic execution of the entire workflow, ensuring the quality of output results effectively.

We have developed a benchmark comprising more than 50 single-cell datasets, comprising dozens of tissues and encompassing hundreds of distinct cell types, encompassing both normal and disease samples. 
Experimental results on 22 datasets with expert-annotated labels demonstrate that CellAgent consistently achieves robust performance, matching or surpassing the performance levels of several best tools. On the entire dataset, CellAgent completes tasks in 92\% of cases, more than doubling the task completion rate compared to direct utilization of GPT-4. 
This innovative approach offers a robust automated solution for data processing and analysis in single-cell transcriptomics research, substantially lowering the technical and financial barriers to entry. Also, CellAgent has the potential to be extensively applied in more biomedical research fields, driving more accurate and comprehensive biological discoveries.

\begin{figure}[t]
\centering
    \includegraphics[width=1\linewidth]{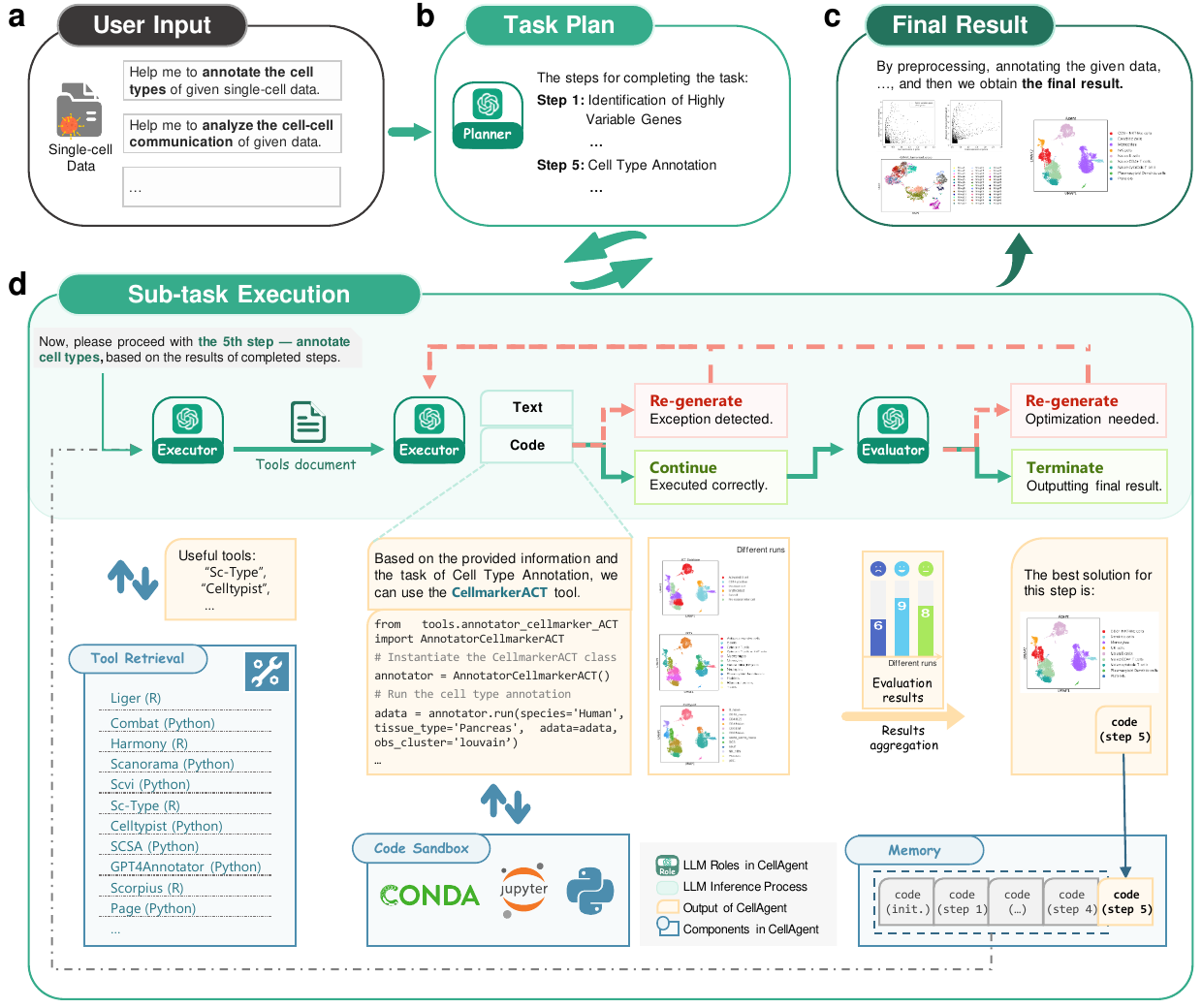}
\caption{\textbf{Schematic of the CellAgent Framework.} \textbf{a}, Example of user input received by the CellAgent, comprising single-cell data and user-provided text information.
\textbf{b}, Upon receiving user input, the Planner role first parses user intent and decomposes the task into subtasks.
\textbf{c}, Illustration of final results, including results of individual subtasks and the final task outcome. 
\textbf{d}, Detailed view of the CellAgent's processing flow for subtasks. The current subtask and historical code memory are inputted to an Executor, which initially retrieves tools and outputs available tools for this step. Subsequently, corresponding documentation for these tools is acquired, and the Executor derives solutions (text analysis and code generation) based on the documentation. These codes are executed in the code sandbox, and if exceptions are encountered, solutions are regenerated until successful execution of this task. Then, the Evaluator assesses the results of the current task and allows the Executor to optimize solutions. Ultimately, based on its evaluation of results under multiple solutions, the Evaluator aggregates results to obtain the final outcome of this step.}
\label{fig:framework}
\end{figure}

\section{Results}
\subsection{A multi-agent collaborative automated workflow for single-cell RNA-seq analysis}

The process of scRNA-seq data analysis is inherently complex and diverse. 
For analyzing a single-cell dataset, tools and parameters are usually required to be manually optimized. The optimal tools and parameters often vary among different datasets.
To address this challenge, we introduce CellAgent, an LLM-driven sophisticated AI agent specialized in automating the entire workflow of scRNA-seq data analysis. 
CellAgent receives data along with a task description in natural language from the user as inputs (Figure~\ref{fig:framework}a).
Subsequently, CellAgent executes a series of automated processes, including task decomposition (Fig.~\ref{fig:framework}b), sequential execution and optimization of subtasks (Fig.~\ref{fig:framework}c), and finally delivers the task results to the user (Fig.~\ref{fig:framework}d).

To enhance the use of LLMs in the context of scRNA-seq data analysis tasks and incorporate domain-specific biological knowledge, we design three distinct LLM-driven biological expert roles, including Planner, Executor and Evaluator. Then, CellAgent effectively organizes the collaboration of these expert roles, through a hierarchical decision-making framework and a self-iterative optimization mechanism. 
Such efficient collaboration enables each role to fulfill its responsibilities, facilitating effective information exchange and collectively achieving high-quality task completion.

\textbf{\textit{Planner:}} To ensure the rational decomposition and step-by-step resolution of complex scRNA-seq data analysis tasks, similar to the process employed by human experts in data analysis, CellAgent introduces a hierarchical decision-making framework. With a precise understanding of the whole workflows related to scRNA-seq data analysis tasks, Planner firstly conducts higher-level planning based on user-provided data and requirements~(Fig.~\ref{fig:framework}b). The Planner possesses the ability to comprehend natural language instructions from the user accurately, identify data features, and design task plans judiciously. This plan includes necessary data preprocessing tasks such as quality control, high variable gene selection, and normalization, as well as potentially required analysis steps such as batch effect correction, cell type annotation, and so on.

\textbf{\textit{Executor:}} Then, as shown in Fig.~\ref{fig:framework}d, the lower-level Executors are responsible for sequentially executing the decomposed subtasks, providing detailed analysis and executable code. Notably, these executors possess a comprehensive understanding of the specific methodologies and tools utilized in scRNA-seq data analysis tasks, and they can acquire documentation for these tools, thereby mitigating code execution failures. If execution exceptions are detected, CellAgent will request Executor to resolve the issue until successful code execution is achieved. 

\textbf{\textit{Evaluator:}} 
Evaluator is responsible for assessing the quality of current results of specific data processing tasks akin to human experts, providing judgments on the effectiveness of various methods. Building upon effective execution of generated code, CellAgent proposes a self-iterative optimization mechanism to ensure the rationality of this solution.
Specifically, based on self-evaluation results or potential code exceptions, Executor automatically optimizes the solutions through hyperparameter tuning or tool selection, leading to iterative improvement of outcomes. Finally, according to the Evaluator's judgment, CellAgent outputs the best solution and corresponding result for this step.

To assess the performance of CellAgent, we conduct detailed evaluations in handling specific tasks: batch effect correction, cell type annotation, and trajectory inference. Our findings reveal that in the majority of cases, the outcomes yielded by CellAgent surpass those generated by other existing tools~(Sec. 2.2-2.4). This elucidates that CellAgent, through effective collaboration among multiple LLM-driven agents, achieves the automation of the scRNA-seq data analysis. The automation process above encompasses tool invocation, code generation, code execution, exception handling, result evaluation, and solution optimization in a closed-loop fashion, culminating in the generation of reliable outcomes. 
To demonstrate the broader applicability of CellAgent across a spectrum of scRNA-seq data analysis tasks, we perform a thorough comparison using various types of datasets.
The results show a remarkable comprehensive completion rate of 92\% for CellAgent, doubling the performance of the GPT-4 model alone. Furthermore, CellAgent also outperforms other scRNA-seq data analysis tools across various tasks.

\begin{figure}[t]
\centering
\includegraphics[width=1\linewidth]{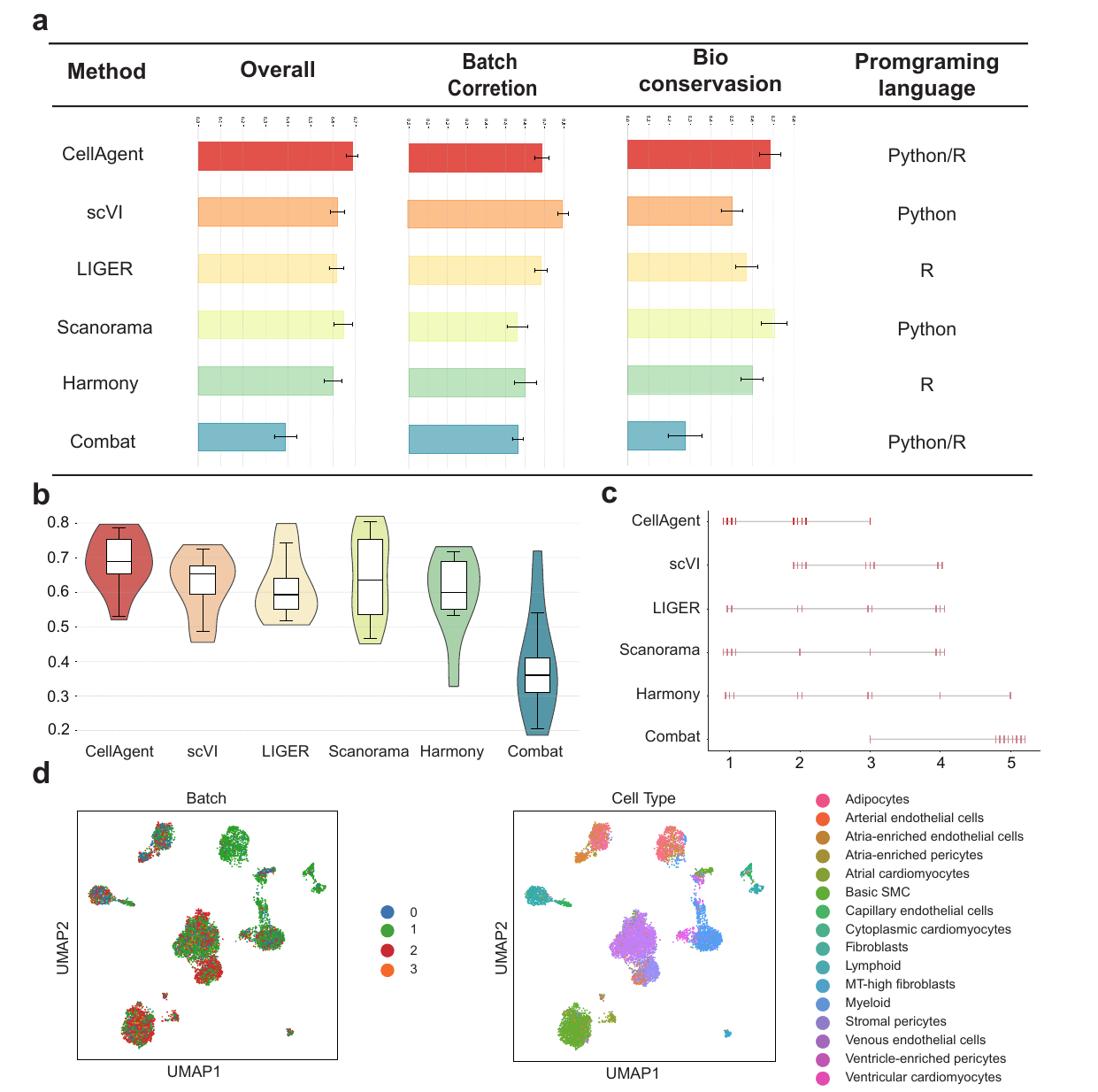}
\caption{\textbf{Batch Correction.} \label{fig:batch}
\textbf{a}, The performance of CellAgent and other batch correction algorithms on batch correction, bio-conservation, and overall scores, along with their programming languages.
\textbf{b}, Violin plot shows the distribution of overall score between CellAgent and other methods across all datasets.
\textbf{c}, The ranking of CellAgent and other methods across different datasets.
\textbf{d}, UMAP plots show the performance of CellAgent on the heart datasets, using batch labels and cell type labels for coloring respectively. 
}
\end{figure}

\subsection{CellAgent enables efficient batch correction}
Batch correction aims to eliminate non-biological variations in datasets from different batches, which stem from differences in experimental conditions, sample handling, or sequencing platforms. These batch effects can obscure or distort the actual biological signals~\cite{batch_benchmark}. Therefore, batch correction is a crucial step in single-cell analysis. 

To evaluate the performance of CellAgent on batch correction, we applied it to nine datasets~\cite{dominguez2022cross}, covering major tissues or organs in the human body. We compared CellAgent with five advanced methods, including scVI~\cite{scVI}, LIGER~\cite{liger}, Scanorama~\cite{Scanorama}, Harmony~\cite{harmony} and Combat~\cite{combat}. Ten metrics are used to evaluate these methods on their ability to remove batch effects while conserving biological variation. Among these metrics, some are for the removal of batch effects (batch correction), such as kBET~\cite{kBET} and $iLISI_{Graph}$~\cite{iLISI-cLISI}, some are for the conservation of biological variance (bio-conservation), such as $cLISI_{Graph}$~\cite{iLISI-cLISI} and ARI~\cite{ARI}. The overall score, derived from averaging ten metrics, serves as the final performance indicator~\cite{batch_benchmark}.

The results indicate that CellAgent achieved top scores in both batch correction and bio-conservation (Fig.~\ref{fig:batch}a). The average overall score of CellAgent reached 0.684 across multiple datasets. CellAgent outperformed the suboptimal method, scVI scoring 0.642. CellAgent achieved the most favorable distribution of overall scores compared with other methods, by observing the quantiles and density estimation curve of the violin plot (Fig.~\ref{fig:batch}b). The overall scores of CellAgent on different datasets are more concentrated, with a median of approximately 0.69, surpassing those of other methods. Additionally, CellAgent ranked first on four datasets, which demonstrated its advantage (Fig.~\ref{fig:batch}c). The UMAP plots show that CellAgent  removed the batch effect while preserving true cell types in the clustering result (Fig.~\ref{fig:batch}d). Cells of the same type, such as ventricle-enriched pericytes cells, myeloid cells, and stromal pericytes cells, from different batches are mixed together.

\begin{figure}[t]%
\centering
\includegraphics[width=1\linewidth]{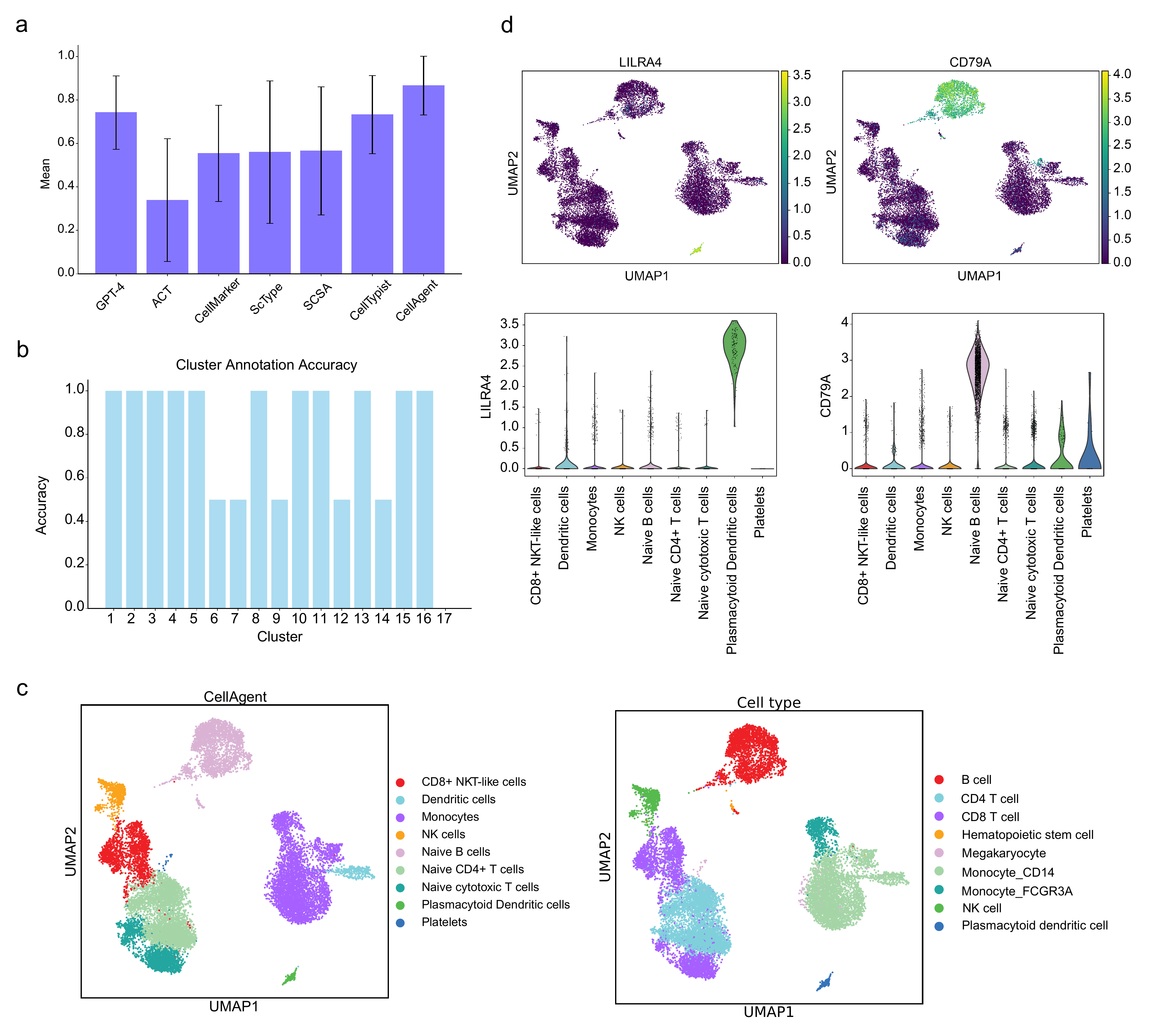}
\caption{\textbf{Cell Type Annotation}\label{fig:celltype}
\textbf{a}, The mean accuracy of cell type annotation results with labeled cell types, derived from samples of five distinct human tissues and mouse tissues.
\textbf{b}, The accuracy of cell type annotation for the 17 clusters of human PBMC dataset.
\textbf{c}, Detailed cell-type annotations of the human PBMC dataset with the CellAgent and expert-annotated cell type Annotations.
\textbf{d}, Visualize the expression of LILR4A gene (PDCs marker) and CD79A gene(B cells marker) across clusters using UMAP plots and violin plots.}

\end{figure}

\subsection{CellAgent facilitates more accurate cell type annotation }
Cell type annotation is a critical yet often time-consuming precursor step in the analysis of scRNA-seq data ~\cite{gpt4anno}. Various tools have been developed for automated cell type annotation. However, these tools exhibit poor generalization capabilities, often performing well only on data from specific tissues or organs while performing poorly in others. 

We conducted benchmarking tests on the performance of seven different methods in terms of their ability to correctly assign cell type annotations.
The datasets used in the benchmarking originated from various tissues, including Human Peripheral Blood Mononuclear Cells~\cite{pbmc}, Human Liver~\cite{liver}, Human Lung~\cite{lung}, Human Pancreas~\cite{pancreas}, and Mouse Retina~\cite{retina}. The diversity of these datasets allowed us to evaluate CellAgent and other methods across different sequencing platforms, tissue types, and organisms. CellAgent showed superior performance in terms of average accuracy across all datasets (Fig.~\ref{fig:celltype}a). 

For example, CellAgent showed high accuracy on the human PMBC dataset, where there exist multiple similar subtypes. We evaluated the consistency between the CellAgent annotations and expert annotations of seventeen clusters in this dataset. Following the evaluation metric in~\cite{gpt4anno}, we categorized the results as `fully match', `partially match', and `mismatch', assigning scores of 1, 0.5, and 0 respectively to each category. The annotation accuracy, revealing that CellAgent achieved `fully match' labeling in ten out of seventeen clusters, `partially match' results in five clusters, and only one cluster with an `mismatch' labeling result. Our model annotated 94\% clusters of the PBMC dataset effectively, achieving a high level of accuracy (Fig.~\ref{fig:celltype}b).
The visualization of the annotation results on the human PBMC dataset, including a comparison with expert-annotated results, also demonstrates the annotations with CellAgent are close to the true cell type labels (Fig.~\ref{fig:celltype}c and Supplementary FigX).

In scRNA-seq data analysis, visualizing the expression of specific genes within each cluster or cell type is crucial. These visualizations are essential for understanding the underlying biological mechanisms. Given a specific genes, CellAgent also provides the function of generating gene expression umap plots and violin plots for users (Fig.~\ref{fig:celltype}d).

\subsection{CellAgent enhances the performance of trajectory inference}

\begin{figure}[t]%
\centering
\includegraphics[width=1\linewidth]{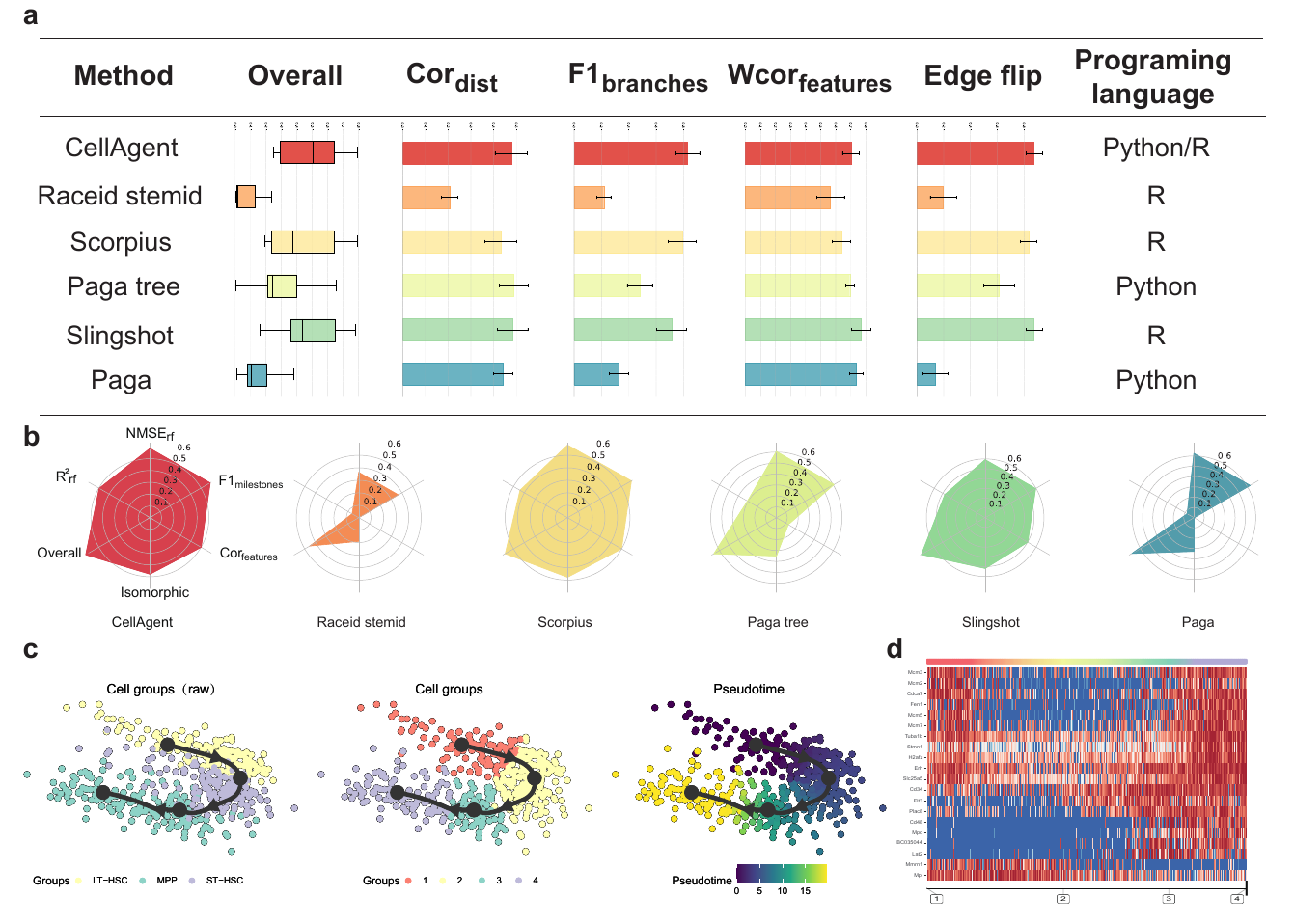}
\caption{\textbf{Trajectory Inference.} \label{fig:trajectory inference}
\textbf{a}, Comparison of the performance between CellAgent and other trajectory inference algorithms on gold standard datasets, along with their programming languages.
\textbf{b}, Radar charts display the performance of CellAgent and other methods on the metrics, which are not covered in a.
\textbf{c}, CellAgent's output on the “aging hsc kowalczyk” dataset.  one UMAP plot depicting trajectories colored by original cell types, another colored by cell type clusters optimized based on milestones within the trajectory, and the last colored by pseudo time.
\textbf{d}, Heatmap of gene expression, with additional emphasis on the expression of milestones within the trajectory. The cell order in the heatmap is optimized according to the trajectory.
}
\end{figure}

Trajectory inference is employed to decode the temporal sequences of cell development and differentiation. It can reconstruct developmental trajectories from static single-cell data, unveil transitions between cell states, and identify changes in various biological properties over time~\cite{saelens2019comparison}. This is crucial for understanding the mechanisms that determine cell fate and the associated complex biological processes. Although new trajectory inference algorithms have emerged in recent years, they are typically suited to specific datasets or topological structures~\cite{todorov2020tinga}. 
Selecting the appropriate trajectory inference algorithm for researchers and discerning its biological significance remain both challenging and valuable.

Through experiments conducted on nine datasets with gold-standard trajectories (see Datasets section)~\cite{saelens2019comparison}. We compared the performance between CellAgent and the other five trajectory inference methods, including Raceid stmid~\cite{RaceID/StemID}, Scorpius~\cite{cannoodt2016scorpius}, Paga, Page tree~\cite{wolf2019paga}, and Slingshot~\cite{street2018slingshot}. Following previous study~\cite{saelens2019comparison},
we utilized $\mathit{cor}_{\textrm{dist}}$,
$\mathit{F1}_{\textit{branches}}$, 
$\mathit{wcor}_{\textrm{features}}$,
$\mathit{edgeflip}$ to measure topological similarity, branch matching, cellular position distances, and correlation of trajectory-specific genes for trajectory evaluation. The weighted average score of these metrics is used to measure the final performance~\cite{saelens2019comparison}.

CellAgent performed best across these datasets with an overall score of 0.496 (Fig.~\ref{fig:trajectory inference}a). It surpassed Slingshot, the suboptimal method at 0.473. Additionally, CellAgent demonstrated  advantages in multiple other metrics~\cite{saelens2019comparison} such as $\mathit{NMSE}_{\textit{rf}}$,
$R^{2}_{rf}$, 
$\textrm{isomorphic}$,
$\mathit{F1}_{\textit{milestones}}$, 
$\mathit{cor}_{\textrm{features}}$
(Fig.~\ref{fig:trajectory inference}b). On the ``aging hsc kowalczyk" dataset, CellAgent revealed the developmental trajectory from LT-HSC cells, through ST-HSC cells, to the final differentiation into MPP cells (Fig.~\ref{fig:trajectory inference}c). CellAgent also depicted the changes in their gene expression patterns (Fig.~\ref{fig:trajectory inference}d). The expression of the CD48 and MPO genes gradually increased during the differentiation process. The accuracy and biological interpretability indicate that CellAgent can help scientists understand the mechanisms determining cell fate and related biological processes.

\section{Discussion}
Single-cell RNA-seq is a powerful tool for characterizing cellular properties, yet it demands high efficiency and quality in advanced data analytics. In this paper, we introduced CellAgent, an LLM-driven automated agent specifically designed for single-cell analysis tasks for the first time, which enables high-quality and automated analysis of scRNA-seq data. 

CellAgent is capable of automatically handling complex single-cell analysis tasks, with its key innovation lying in the effective collaboration among multiple bio-expert LLM roles. This automation capability stems from its use of powerful AI generative models, which are capable of understanding human natural language instructions and generating corresponding content. Notably, we found that GPT-4, a state-of-the-art LLM, has a foundational understanding of specific biological knowledge, including common gene markers and various single-cell analysis tasks. However, its limitations are evident in its propensity to confuse certain specialized concepts. To address this, CellAgent expands the LLM's capability boundaries through a manually defined tool library and several experiential knowledge, thereby enhancing its proficiency in handling single-cell analysis tasks. Then, we create different bio-expert roles within CellAgent based on GPT-4, enabling each role to focus on solving specific processes. 
Subsequently, CellAgent enhances the efficiency of addressing complex tasks by connecting the task planner with various roles responsible for task execution through hierarchical planning. Also, by employing a self-iterative optimization mechanism and utilizing modules such as Tool retriever, Memory module, and Code sandbox, CellAgent further enhances the quality of output results. 

It is also worth mentioning that CellAgent proposes using GPT-4V to automatically evaluate the results of batch correction and trajectory inference, guiding the self-iterative optimization process. To our knowledge, this is the first time a multi-modal LLM~(MLLM) like GPT-4V has been applied to evaluate these results, as a replacement for human judgment. Additionally, CellAgent proposes the first attempt of using GPT-4 to effectively aggregate the results from cell type annotation results of different tools, obtaining a more reasonable final annotation. These developments provide new insights into applying LLMs (or MLLMs) to bioinformatics research.

Overall, CellAgent is a versatile, extensible, and automated tool for scRNA-seq data analysis. Its process of task completion does not rely on human intervention, greatly reducing the difficulty and cost of data analysis. Also, its open architecture enables users to provide specific new knowledge and tools, allowing the CellAgent to better align with user expectations, as an ideal assistant for researchers. The emergence of CellAgent not only opens up new research directions in bioinformatics but also expands the application of generative AI in sciences, leading to potential new findings and a deeper understanding of biological systems.

Testing on several datasets with expert-annotated labels revealed that the results obtained by CellAgent are, in most cases, the best among all tools, demonstrating the powerful ability of CellAgent to automate scRNA-seq data analysis and ensure quality, reaching the level of human experts. The overall evaluation indicates that CellAgent significantly outperforms the data analysis process involving only a single GPT-4 model, illustrating that this multi-agent collaboration framework deeply taps into the potential of GPT-4 in handling single-cell data analysis tasks.

However, a notable limitation lies in the imperfect self-evaluation methods of the CellAgent. Presently, self-evaluation primarily relies on GPT-4V or GPT-4. While these innovative evaluation processes contribute to self-optimization and ensure final effectiveness, CellAgent is limited to support a more diverse range of optimization directions. Therefore, when users have specific preferences regarding optimization goals, manual specification of evaluation processes can prompt CellAgent to optimize according to user expectations. Nevertheless, enabling CellAgent to efficiently and automatically integrate new tools provided by users to align with user requirements represents a valuable research direction for the next stage of development in AI agents for science.

\section{Methods}
\subsection{The composition of CellAgent}
Facing scRNA-seq data analysis tasks, CellAgent enables automated task decomposition, code execution, and iterative optimization. During this process, three LLM-driven biological expert roles and their communication and collaboration mechanisms serve as the core brain of CellAgent, responsible for thinking and providing solutions. Besides, to achieve efficient automation, auxiliary components such as memory~$\mathcal{M}$, tool retrieval~$\mathcal{T}$, and code sandbox~$\mathcal{E}$ are essential. These components enable CellAgent to effectively manage and execute tasks, allowing the three biological expert roles to focus on their thinking and providing efficient solutions, ensuring seamless automation of the whole workflow.

\subsubsection{Biological Expert Agents}
CellAgent comprises three distinct roles of biological expert LLMs with different responsibilities, including Planner, Executor, and Evaluator. To distinguish these roles and enhance task execution efficiency, we collect a small amount of human expert experience about scRNA-seq data analysis in natural language and integrate necessary tools, then inform different LLMs about role assignments, specific expert experiences, and tools, thus constructing these biological expert roles. Notably, the user input of CellAgent includes three aspects: the data to be processed (${D}$), the description of the task to be solved ($u_\text{task}$), as well as the optional data description ($u_{D}$) the preference requirements for solving the problem or tools ($u_\text{req}$).

\paragraph{Planner} Planner is designed to understand user requirements and data, and provide comprehensive task planning. Therefore, the system prompt for the Planner ($p^p_\text{sys}$) primarily includes 1) role positioning descriptions, detailing role inputs and outputs; 2) output format specifications for task planning, with the Planner's output format set to JSON to facilitate subtask extraction; and 3) pre-collected expert experience information.
Upon receiving user input, the Planner is also informed of the representation parsed from the data to be processed, making it easier to understand data features. Specifically, we use \verb|AnnData| in Python to process single-cell data, and obtain its string representation, denoted as $\psi(D)$. We denote the LLM role Planner as $\mathcal{A}^{\text{LLM}}_p$, then the process of generating the description of $n$ sub-steps according these information is:
\begin{equation}
    \{t_1, t_2, ..., t_n\} \leftarrow \mathcal{A}^{\text{LLM}}_p(p^p_\text{sys}, u_\text{task}, u_\text{req}, u_D, \psi(D)).
\end{equation}

\paragraph{Executor} Executor agents are included in CellAgent to enhance the efficiency of specific task execution, with the \textbf{Tool Selector} solely responsible for selecting available tools, and \textbf{Code Programmer} responsible for generating code to complete the current task based on the tools recommended. 

The system prompt for the Tool Selector, denoted as $p^t_\text{sys}$, mainly includes 1) role positioning descriptions with detailed input and output specifications; 2) output format specifications for ease of use by other roles, with JSON specified as the output format. 
During execution of the $i$-th step, Tool Selector first retrieves the list of all tools $\mathcal{T}$ and the description of the current step $t_i$. Additionally, to ensure alignment with the user's possible needs at the whole procedure, its prompt also includes user requirement. We denote the Tool Selector as $\mathcal{A}^{\text{LLM}}_t$, which generate a list of available tools in this specific steps, represented as: 
\begin{equation}
    \mathcal{T}_{t_i} \leftarrow \mathcal{A}^{\text{LLM}}_t(p^t_\text{sys}, u_\text{req}, \mathcal{T}, t_i).
\end{equation}

The system prompt for the Code Programmer ($p^c_\text{sys}$) mainly includes 1) role positioning descriptions that detail the inputs and outputs; 2) output format specifications that clearly define the format of output code blocks for easy parsing and execution; and 3) a few human experience, about professional tools usage and common exceptions resolution.
During execution, it automatically retrieves the documentation of the tools available in this step $(\text{Doc}(\mathcal{T}_{t_i}))$, and is also provided with the task description for the current step, data representation, and user preference requirements. 
Besides, memory about history code for previous steps is prompted, denoted as $\mathcal{M}$, enabling the code generation process to better match the context.
Specifically, we denote the Code Programmer as $\mathcal{A}^{\text{LLM}}_c$, and its response contains text analysis $w_i$ and the code $c_i$ for this specific step $t_i$, as:
\begin{equation}
    (c_i, w_i) \leftarrow \mathcal{A}^{\text{LLM}}_c(p^c_\text{sys}, u_\text{req}, u_D, \mathcal{M}, t_i, \text{Doc}(\mathcal{T}_{t_i})).
\end{equation}
Then, the code $c_i$ is executed in Code Sandbox. The execution results, denoted as $\mathcal{E}(c_i)$, can be potential execution exception, or the successfully executed results. If exceptions occur, the Code Programmer executes fixing actions to generate the rectified code, as $\mathcal{A}^{\text{LLM}}_t(\mathcal{E}(c_i))$.

\paragraph{Evaluator} Evaluator is tasked with assessing the results of the current step and choosing the best among the multiple outcomes produced by the Executor's self-optimization. The system prompt for the Evaluator ($p^e_\text{sys}$) mainly includes 1) role positioning descriptions, including input and output descriptions; 2) a list of available evaluation methods, including specific codes integrated into CellAgent for evaluating key steps. During execution, it receives the string representation of data, the task description for the current step, user preference requirements, and most crucially, the execution codes. Subsequently, the Evaluator conducts an evaluation. If in current trial, the Evaluator can assess the results of multiple trials and select the optimal solution, the final solution for the current step will be determined. Otherwise, the Code Programmer will be prompted to optimize the solution.  We denote the best solution code for step $t_i$ as $\bar{c_i}$, and denote the code for the several trails as $\{c_i^j, j=1, 2, ...\}$. Then, the process of the Evaluator evaluating and selecting the optimal solution code for the current step can be represented as:
\begin{equation}
    \bar{c_i} = \mathcal{A}^{\text{LLM}}_e(p^e_\text{sys}, u_\text{req}, u_D, t_i, \{c_i^j\}), j=1, 2, ... .
\end{equation}

\subsubsection{Auxiliary components}
\paragraph{Memory Control}
LLMs do not have the capability to retain historical messages, which necessitates incorporating a memory module in LLM-driven Agents. By organizing past information rationally through the memory module, LLMs can prevent forgetting previous content during planning and process long contexts more efficiently. In CellAgent, the internal logic of different subtask processing is mutually independent, meaning that each subtask execution process does not need to perceive the internal processing of previous steps, such as tool selection, code exception handling, or intermediate code generated during iteration. Therefore, CellAgent defines both global memory and local memory, storing historical information efficiently. This enables different LLM roles to access proper valuable historical information, boosting LLM inference efficiency.

As shown in Fig.~\ref{fig:framework}d, we define global memory to store only the final code of each historical step, which can be represented as $\mathcal{M} \leftarrow \{\bar{c_1}, \bar{c_2}, ...  \}$. This allows the Executor to generate new code more effectively, reducing redundant work, and increasing code generation accuracy. We use code as a memory storage medium, which has high information entropy in scRNA-seq data analysis, enabling the transfer of comprehensive information with fewer tokens, reducing LLM costs, and improving LLM efficiency.

Within each subtask processing, the Executor maintains a local memory that only retains dialogue information within the current step and resets it when the subtask ends. The local memory enables the Executor to perceive the entire self-optimization process, including each generated correct or incorrect code snippet, potential error messages, and optimization processes. This helps avoid repeated errors during optimization, improving processing efficiency.

\paragraph{Tool Retrieval}
CellAgent integrates multiple tools for single-cell analysis tasks to ensure operational stability. This integration is primarily facilitated by the Tool Retrieval module, denoted as $\mathcal{T}$. The integrated tools are registered within the CellAgent framework, allowing the Tool Selector to detect their presence and retrieve a list of potentially useful tools for Code Programmer at the beginning of each subtask. Additionally, in our implementation, the Tool classes are equipped with standardized documentation, known as docstrings in Python. This feature enables the Executor to access documentation for the selected tools, enhancing the accuracy of code generation.

\paragraph{Code Sandbox}
To ensure the security and reliability of code execution, CellAgent implements a Code Sandbox, isolating the code generated by LLMs for execution. Specifically, this is achieved  through Jupyter Notebook Conversion (\verb|nbconvert|), wherein data loading and each step of code generated by LLMs are executed within a comprehensive Jupyter notebook. This implementation approach decouples the CellAgent framework's running and code execution of single-cell data analysis, enhancing the security of executing generated code. Additionally, it facilitates result management for single-cell task analysis tasks and reproducibility.

\subsection{Self-iterative Optimization via Self-evaluation}

\subsubsection{Preprocessing} 
CellAgent first processes the data by removing low-quality cells and genes,  subsequently identifies highly variable genes. Then, the Evaluator checks the effectiveness of these preprocessing steps, optimizes the preprocessing code to fit the characteristics of the dataset better, and showing proper visual plots to users. Unless specified by the user, CellAgent is required to execute a single iteration by default of self-optimization for preprocessing step.

\subsubsection{Batch Correction}
CellAgent utilizes an iterative optimization approach to produce the best batch correction result by assessing the outcomes of different methods through UMAP plots. Specifically, the Evaluator, integrated with GPT-4v, evaluates these plots and ranks these methods. By default, the iteration count is set to three, indicating that the self-optimization process will yield three distinct batch correction results. Following evaluation, the method with the highest score will be selected as the final output.

\subsubsection{Cell Type Annotation}
CellAgent employs both database-based annotation tools, such as CellMarker 2.0 and ACT, and gene expression-based tools, including CellTypist, SCSA, ScType, and GPT-4, for cell type annotation. Utilizing a self-iterative optimization mechanism, CellAgent autonomously selects various methods to execute and obtain results. The Evaluator, driven by GPT-4, then aggregates these annotations and determines the final cell type for each cluster.

\subsubsection{Trajectory Inference}
The evaluation of trajectory inference in CellAgent is analogous to that of batch correction. As trajectory information is challenging to describe directly through text, the Evaluator scores the visualization results of various trajectory inference algorithms. It then selects the highest-scoring visualization as the final output during the iterative optimization process. These visualizations typically consist of UMAP plots that depict trajectories, integrating both cell and trajectory information to enrich the content. Additionally, expert-written knowledge about trajectory inference tasks is provided to the Evaluator in the form of prompts.

\subsection{Evaluation metrics}

\subsubsection{Cell Type Annotation Metrics}To assess the accuracy of cell type annotations and facilitate comparisons across various methodologies, we employ an average scoring metric. For predicted and actual labels, 'fully match' is awarded when they directly align in terms of annotation terms or CL cell ontology names. 'Partial match' occurs when labels share general cell type category (e.g., fibroblast and stromal cell) but differ in specific annotations or ontology names. Conversely, 'mismatch' is declared when there is discrepancy in broad cell type categories, annotations, or ontology names. We assign consistency scores of 1, 0.5, and 0 for 'fully match', 'partial match', and 'mismatch', respectively~\cite{gpt4anno}. These scores are then averaged across all cell types within each dataset. This average score serves as quantitative measure of annotation accuracy, enabling robust comparisons between different annotation methods and datasets. This systematic scoring approach ensures that researchers can objectively assess precision of cell type classifications and refine their analytical methods or data interpretation as needed. 

\subsubsection{Batch Correction Metrics}
Ten different metrics are employed to assess the capacity to eliminate batch effects while preserving biological variation. These metrics are divided into two categories~\cite{batch_benchmark}. The metrics representing the effectiveness of batch effect removal include Graph Connectivity,  PCR Comparison~\cite{batch_benchmark}, $iLISI_{Graph}$~\cite{iLISI-cLISI}, kBET~\cite{kBET}, and $ASW_{batch}$~\cite{Silhouette}. Metrics representing bio-conservation are Isolated Labels~\cite{batch_benchmark}, ARI~\cite{ARI}, NMI~\cite{NMI}, $ASW_{cell}$~\cite{Silhouette}, and $cLISI_{Graph}$~\cite{batch_benchmark}. The weights of these metrics respectively represent scores of batch correction and biological conservation. The overall score is the weighted average of all these metrics as the final indicator.

\subsubsection{Trajectory Inference Metrics}
Four distinct types of metrics are employed to evaluate different aspects of trajectories. These metrics include $\mathit{edgeflip}$  measuring the similarity between two topologies,  
$\mathit{F1}_{\textit{branches}}$ and $\mathit{F1}_{\textit{milestones}}$ assessing the similarity in the assignment of cells onto branches,  $\mathit{cor}_{\textrm{dist}}$ assessing the similarity in cellular positions between two trajectories~\cite{saelens2019comparison}. Furthermore, metrics such as $\mathit{wcor}_{\textrm{features}}$ and $\mathit{cor}_{\textrm{features}}$ quantify the agreement between differentially expressed features from the known trajectory and the predicted trajectory. The overall score is calculated as the geometric mean of $\mathit{edgeflip}$, $\mathit{F1}_{\textit{branches}}$, $\mathit{cor}_{\textrm{dist}}$ and $\mathit{cor}_{\textrm{features}}$ serving as the comprehensive trajectory evaluation metric~\cite{saelens2019comparison}.

\section{Data availability}
For the batch correction task, we utilized nine datasets from different human organ atlases including blood, bone marrow, heart, intestine, kidney, liver, lung, pancreas, and skeletal muscle. They are available at \url{https://www.celltypist.org/organs}. We employed the following datasets for the cell type annotation task: PBMC dataset (\url{https://www.10xgenomics.com/resources/datasets/8-k-pbm-cs-from-a-healthy-donor-2-standard-2-1-0}); pancreas dataset including Smart-seq2, 10X Chromium, and Drop-seq (\url{https://hemberg-lab.github.io/scRNA.seq.datasets/human/pancreas/}); multiple PBMC dataset can be obtained from Single Cell Portal and the accession code is SCP424; multi-Omics (\url{https://figshare.com/projects/Tabula_Muris_Transcriptomic_characterization_of_20_organs_and_tissues_from_Mus_musculus_at_single_cell_resolution/27733}); immune dataset from different donors (\url{https://figshare.com/ndownloader/files/25717328}); mouse dataset from the brain region (\url{https://portal.brain-map.org/atlases-and-data/rnaseq/mouse-whole-cortex-and-hippocampus-10x}). We employed the following datasets for the trajectory inference task: the aging-hsc-old Kowalczyk, and aging-hsc-young Kowalczyk datasets, which encompass developmental trajectories of human hematopoietic stem cells; human-embryos petropoulos dataset, containing developmental trajectories of human embryonic cells; NKT-differentiation engel dataset from thymus; pancreatic-alpha-cell-maturation and cellbench-SC1, germline-human, cell-cycle, and stimulated-dendritic-cells-PIC datasets. These datasets are deposited on Zenodo (\url{https://doi.org/10.5281/zenodo.1443566}).

\bibliography{main}

\end{document}